\documentclass{article}

\usepackage{microtype}
\usepackage{graphicx}
\usepackage{subfigure}
\usepackage{booktabs} 
\usepackage{subcaption}
\usepackage{subfiles}
\usepackage{xcolor}
\usepackage{array}
\usepackage{amsmath} 
\usepackage{listings}
\usepackage[scaled=0.82]{beramono} 
\usepackage{stfloats}

\newcolumntype{L}[1]{>{\raggedright\arraybackslash}p{#1}}

\usepackage{hyperref}

\newcommand{\bugs}{142}
\newcommand{\pendingbugs}{103}

\newcommand{\squishlist}{
   \begin{list}{$\bullet$}
    { \setlength{\itemsep}{0pt}      \setlength{\parsep}{3pt}
      \setlength{\topsep}{3pt}       \setlength{\partopsep}{0pt}
      \setlength{\leftmargin}{1em} \setlength{\labelwidth}{1em}
      \setlength{\labelsep}{0.5em} } }

\newcommand{\squishend}{
    \end{list}  }     

\usepackage[accepted]{mlsys2025}

\mlsystitlerunning{XLA from the Trenches: Correctness on TPU Workloads at Scale}

\begin{document}

\twocolumn[
\mlsystitle{XLA from the Trenches: Correctness on TPU Workloads at Scale}



\mlsyssetsymbol{equal}{*}

\begin{mlsysauthorlist}
\mlsysauthor{Ignacio Cano}{goo}
\mlsysauthor{Amir Saadat}{goo}
\mlsysauthor{Rishipal Singh Bhatia}{goo}
\mlsysauthor{Samuel Ginzburg}{goo}
\mlsysauthor{Patrick C.~Toulme}{goo}
\mlsysauthor{Ioannis Zarkadas}{goo}
\mlsysauthor{Praveen Narayanan}{goo}
\mlsysauthor{Pratik Bhatu}{goo}
\mlsysauthor{Nicolas Basile}{goo}
\mlsysauthor{Alexander Dietmüller}{goo}
\mlsysauthor{Ruishan Wang}{goo}
\mlsysauthor{Jacob Wahlgren}{goo}
\mlsysauthor{Victor Barros}{goo}
\mlsysauthor{Subhankar Shah}{goo}
\mlsysauthor{Michael Kuperstein}{goo}
\mlsysauthor{Farzin Houshmand}{goo}
\mlsysauthor{Emma Sevastian}{goo}
\mlsysauthor{Peter Gavin}{goo}
\mlsysauthor{Karthik Srinivasa Murthy}{goo}
\mlsysauthor{Berkin Ilbeyi}{goo}
\mlsysauthor{Amit Sabne}{goo}
\mlsysauthor{Eli Bendersky}{goo}
\end{mlsysauthorlist}

\mlsysaffiliation{goo}{Google}

\mlsyscorrespondingauthor{Ignacio Cano}{nachocano@google.com}


\vskip 0.2in

\begin{abstract}

Accelerating modern foundational models has pushed domain-specific compilers toward increasingly complex, multi-tiered optimization pipelines. However, this heavy optimization focus introduces delicate tradeoffs between peak execution performance and correctness.
Subtle codegen defects can trigger numerical degradation and memory or buffer corruption, invalidating multi-million-dollar training runs and degrading live serving quality without raising obvious compiler errors.
In this paper, we present a battle-tested methodology for verifying and triaging Google's XLA:TPU compiler at hyperscale. We leverage fuzzing techniques, combining synthetic program generation with real production traces, while orchestrating AI agent swarms to autonomously generate stress tests, reproduce failures, and draft fixes. To pinpoint root causes in full workloads, we pair this with a hierarchical differential pipeline that bisects large graphs down to individual failing operations inside fused kernels. Deployed in production, our framework has already driven the resolution of over~\bugs{} compiler bugs (with~\pendingbugs{} additional findings under review), from which we distill pragmatic lessons learned.


\end{abstract}

]



\printAffiliationsAndNotice{}  

\section{Introduction}


Optimizing machine learning (ML) workloads has become one of the defining computational challenges of our era. At the center of this revolution lie domain-specific ML compilers, such as XLA~\cite{xla}, TVM~\cite{chen2018tvm}, Triton~\cite{Tillet2019Triton}, Glow~\cite{DBLP:journals/corr/abs-1805-00907}, which bridge the gap between high-level framework representations and custom accelerator architectures like Google's Tensor Processing Units (TPUs)~\cite{tpus2017} and modern GPUs. To unlock the physical compute limits of hardware, modern ML compilers rely heavily on aggressive, hardware-specific optimization passes. Crucial to this process are tiled fusions, where the compiler merges multiple operations into unified execution kernels and partitions them into smaller tiles, streaming intermediate data through fast on-chip caches or scratchpad memory rather than materializing it in off-chip High Bandwidth Memory (HBM).

While substantial literature and industrial efforts in ML compilation focus on performance—maximizing FLOP utilization, minimizing memory footprint, and automating tuning heuristics~\cite{adams2019halide, chen2018tvm, chen2018learning, cano2026catwild}—an equally vital requirement is compiler correctness. In the era of frontier foundational models and safety-critical autonomous systems, a compiler defect can have severe, widespread repercussions. 
Beyond explicit failures like segmentation faults or fatal assertions, ML compiler bugs frequently manifest as silent numerical drift or subtle data corruption.
A single miscompiled index offset, improper loop accumulator revisit, or packing error can silently alter tensor outputs, causing multi-million-dollar training runs to diverge after weeks of compute, or injecting subtle hallucinations into live inference services~\cite{anthropic2025postmortem, roune2025testing}.


Guaranteeing correctness in a production ML compiler faces several distinct challenges. First, code generation for accelerators involves a combinatorial explosion of interacting knobs (e.g., tensor shapes, tile sizes, buffering levels, layouts) creating an optimization space where standard regression suites barely scratch the surface. Second, modern production workloads compile into large computational graphs, making whole-model triage of numerical divergence a significant bottleneck that can take weeks of manual toil. Third, while differential testing against an unoptimized reference interpreter is theoretically sound, naive synthetic inputs trigger undefined floating-point math and hardware alignment violations, generating false-positive noise that undermines automated isolation. Finally, even after isolating an anomaly to a single fusion node, pipelined kernels execute intermediate values purely in hardware vector registers (without materializing them in HBM), turning fusions into opaque black boxes that blind standard debugging tools.

To address these challenges, we introduce a cohesive reliability framework for the XLA:TPU compiler that tackles correctness both proactively and reactively. Proactively, we improve bug detection via hybrid fuzzing. Rather than relying solely on (unrepresentative) synthetic graphs, we piggyback differential correctness checks onto our existing production autotuning pipeline~\cite{cano2026catwild}, stress-testing boundary conditions across real fleet operations at near-zero marginal infrastructure cost. To further amplify this approach, we leverage AI to systematically expand test coverage across the fleet and scale both the findings and the generation of compiler fixes. Reactively, we establish a hierarchical differential isolation pipeline spanning from macro-graphs down to fused kernels. At the graph level, we recursively navigate large computational graphs to pinpoint failing operations while reducing false positives using reverse constraint propagation. At the kernel level, we inspect intermediate tensors directly inside fused kernels to pinpoint the exact sub-operation where hardware execution diverges from the reference semantics.




In summary, our main contributions are:

\squishlist
\item We design and deploy a hybrid fuzzer that evolved from purely synthetic XLA program generation, to leveraging real fleet-mined operations via production autotuning sweeps, and ultimately to orchestrating autonomous agent swarms that scale our reliability impact across the fleet.
\item We present a unified diagnostic framework spanning from macro-graph navigation to micro-kernel inspection, combining constraint-propagated input generation to isolate failing subgraphs with intra-fusion intermediate inspection to localize defects inside opaque fused kernels.
\item We evaluate our methodology on real-world production workloads---resolving over~\bugs{} compiler bugs to date, with~\pendingbugs{} additional fixes under review---and distill pragmatic lessons learned from testing the XLA:TPU compiler at datacenter scale.
\squishend


\section{Background \& Motivation}

We conduct our work within the context of Google's ML production fleet, targeting the XLA compiler across multiple generations of TPUs. In this section, we first outline the software and hardware architectural foundations, focusing on memory hierarchy constraints and tiled operator fusion, that make accelerator code generation uniquely challenging. Next, we describe how compiler defects manifest in production. Finally, we present an empirical analysis of our fleet testing challenges, illustrating the widening gap between traditional regression suites and the combinatorial reality of production workloads.

\subsection{XLA:TPU Overview}

XLA:TPU lowers ML programs through HLO $\rightarrow$ LLO $\rightarrow$ VLIW bundles. The High-Level Optimizer (HLO) intermediate representation (IR) encodes whole-tensor dataflow graphs where global passes perform optimizations like algebraic simplification, layout assignment, and operator fusion. Fused subgraphs then lower to the Low-Level Optimizer (LLO) IR—modeling tile loops, DMAs, and compute units (MXU, VPU)—where backend passes apply software pipelining, loop peeling, and register allocation to emit VLIW bundles.

TPUs pair dense TensorCores (and SparseCores, when present) with a compiler-managed memory hierarchy---off-chip High Bandwidth Memory (HBM) and a fast on-chip SRAM scratchpad (Vector Memory, or VMEM)---feeding $8\times 128$ (sublane $\times$ lane) Vector Registers (VREGs), also referred to as chunks. 
Because HBM has lower bandwidth than VMEM and TensorCore compute units operate directly on VREGs, the compiler explicitly stages data from HBM into VMEM via DMA before loading sub-tiles into VREGs.

Tiled operator fusion (windowed execution) is XLA:TPU's cornerstone optimization, serving as the compiler's primary workhorse for eliminating costly HBM round-trips. By merging producer-consumer chains and slicing tensors into VMEM-resident windows, tiles are pipelined from HBM into VMEM, streamed across the fused ops in VREGs, and written back to HBM only once.

\subsection{Anatomy of Miscompilations on TPUs}
\label{sec:anatomy}

\paragraph{Compile-Time Crashes} Compiler failures in XLA:TPU span three levels of difficulty, beginning with \emph{compile-time crashes} (internal compiler errors), such as assertion failures in HLO or LLO passes, unhandled tensor layouts, or VMEM allocation exhaustion. Strictly speaking, these errors are not miscompilations because the compiler aborts before emitting an executable.
While disruptive, they are the easiest class of bugs to catch and triage, they require no hardware execution or correctness oracle, and the compiler fails loudly with a stack trace pointing directly to the offending pass.

\paragraph{Runtime Crashes} True miscompilations occur when the compiler succeeds but generates flawed machine code, starting with runtime crashes and hangs. Here, the compiled binary launches on the TPU but violates hardware invariants (e.g., issuing out-of-bounds HBM-to-VMEM DMA addresses or omitting LLO synchronization barriers, which triggers a device exception or TensorCore timeout). These bugs represent a medium level of difficulty; although the failure is decoupled from compilation and requires real TPU execution to observe, the hardware still provides an unmistakable, fail-stop error signal.

\paragraph{Silent Numerical Miscompilations} The most difficult failures---and our primary focus---are silent numerical miscompilations, where the binary compiles and executes to completion without error but produces corrupted tensor outputs.
On TPUs, these deterministic bugs often arise from subtle corner cases in tiled operator fusion and LLO lowering, including unsound HLO algebraic rewrites, off-by-one bounds during window calculations, multi-buffered DMA race conditions in VMEM, or VREG clobbering across VLIW bundles. Because silent corruptions emit no failure signal and masquerade in production as ML training instability or degraded model quality, catching them is uniquely challenging and requires a differential oracle capable of separating true compiler bugs from benign floating-point non-associativity. A closely related (and equally challenging) class is non-determinism bugs, where asynchronous DMA race conditions in multi-buffered VMEM pipelines or reads from uninitialized VMEM/VREG padding cause run-to-run output divergence on identical inputs; while our methodology can occasionally surface these flaws, they are not our primary focus.

\subsection{Limitations with Traditional Testing Approaches}




One might expect continuous integration (CI) to catch compiler regressions before they reach production. Indeed, the XLA compiler is guarded by an extensive testing pipeline. The source code contains more than 40k static unit test definitions. Parameterization over data types (e.g., \texttt{fp32}, \texttt{bf16}, \texttt{fp8}, \texttt{int8}), shapes, and layouts expands these into nearly 200k distinct test cases. Each test case is then run on every backend it supports, from CPU-only builds and simulators to on-device execution on several TPU  generations. Based on a random sample of test runs, a single pass over the full suite amounts to well over a million test case executions.

Despite this scale, silent miscompilations continue to escape due to two fundamental limitations: (1) Combinatorial explosion of the compilation space: the space of valid operator fusions across arbitrary production graphs is vast, making exhaustive static coverage impossible. Beyond fusion formation itself, downstream backend parameters---such as tiles/windows sizes, multi-buffering levels, and LLO flags---multiplicatively explode the state space, leaving complex corner cases untested. (2) Prohibitive cost of long-horizon training validation: the most insidious numerical miscompilations do not trigger immediate failures, but instead manifest subtly over long-running workload training (e.g., degraded convergence velocity or late-stage loss instability), which is prohibitively expensive to execute as part of regular CI testing.

\begin{figure}[h!]
    \centering
    \includegraphics[width=0.8\linewidth]{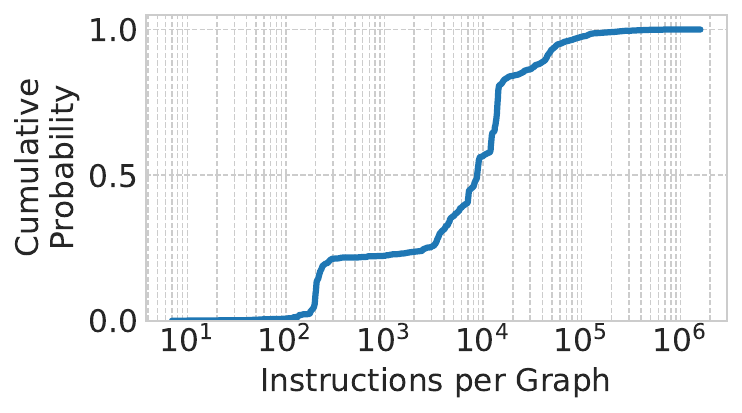}
    \caption{CDF of Graph Instructions.}
    \label{fig:graph_sizes}
\end{figure}

When a numerical divergence or failure surfaces in production, the immediate priority is rapid triage and fault localization. However, pinpointing a defective operation(s) within computational graphs of tens of thousands of nodes is intractable without proper tooling. To illustrate this scale, Figure~\ref{fig:graph_sizes} plots the cumulative distribution function (CDF) of instruction counts across a representative sample of around 200k production graphs executing on a particular TPU generation. More than 40\% of the fleet graphs in this sample exceed 10k nodes, even when counting each fused kernel as a single collapsed node rather than unrolling its internal operations.


\begin{figure}[h!]
    \centering
    \includegraphics[width=0.8\linewidth]{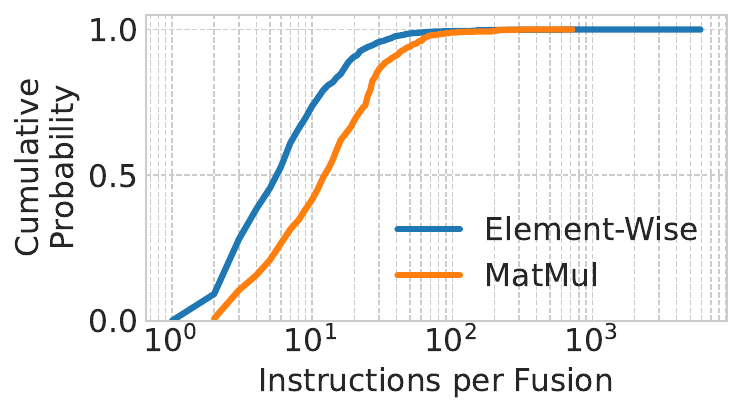}
    \caption{CDF of Fusion Instructions.}
    \label{fig:fusion_sizes}
\end{figure}

Isolating a failure to a single collapsed fusion node is only half the battle, as fused subgraphs themselves can be substantial. Figure~\ref{fig:fusion_sizes} plots the CDF of internal instruction counts per fusion from the same fleet sample, categorized into element-wise fusions (comprising purely element-wise operations such as additions) and matmul fusions (containing at least one matrix multiplication). While element-wise fusions are generally smaller in the median than matmul fusions---where 25\% exceed 25 nodes (p75=25)---both categories exhibit heavy tails reaching hundreds of instructions for matmul fusions (up to $\sim$700) and thousands for element-wise fusions (up to 5,900). Because intermediate values inside a fusion remain resident in on-chip VMEM and VREGs rather than materializing in HBM, manually narrowing down a root-cause defect across even a few dozen (let alone hundreds) of fused nodes is intractable, requiring automated intra-fusion inspection.


\section{Methodology}

\begin{figure*}[t]
\centering
\includegraphics[width=\textwidth]{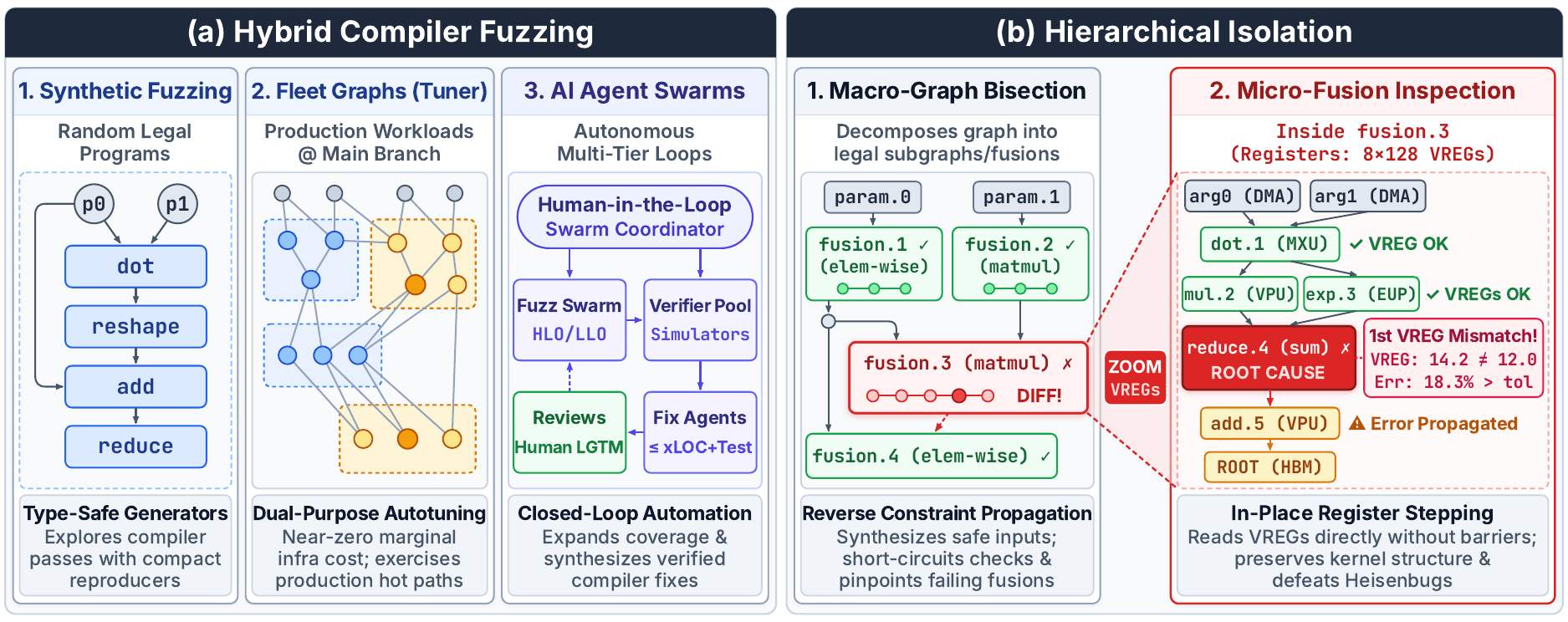}
\caption{Methodology Overview: Hybrid Compiler Fuzzing and Hierarchical Isolation.}
\label{fig:unified_methodology}
\end{figure*}


In this section, we present our two-pronged methodology (Figure~\ref{fig:unified_methodology}): hybrid compiler fuzzing to uncover and fix latent miscompilations (Section~\ref{sec:fuzzing}), and multi-scale hierarchical isolation to pinpoint root-cause operations inside large-scale production graphs and monolithic fused kernels (Section~\ref{sec:isolation}).

\subsection{Hybrid Compiler Fuzzing}
\label{sec:fuzzing}


To overcome the combinatorial and observability limits of static CI suites, our proactive fuzzing strategy evolved across three stages: from constrained synthetic program generation, to opportunistic fuzzing of real fleet workloads via production autotuning, and ultimately to autonomous AI agent swarms that scale both coverage and patch synthesis.

\subsubsection{Template-Guided Synthetic Program Generation}
\label{sec:xlasmith}


Hand-written unit tests inherently suffer from developer prior bias; that is, compiler engineers primarily test corner cases they already anticipate when authoring a pass. To systematically explore unforeseen interactions across passes, we first developed a template-guided synthetic program generator inspired by Csmith~\cite{yang2011finding} and QuickCheck~\cite{10.1145/357766.351266}. Rather than randomly mutating existing graphs (which frequently invalidates structural invariants) our generator constructs syntactically valid programs from scratch via bottom-up, recursive, type-directed generation.

Concretely, generation operates over typed functional intermediate representations (Jaxpr\footnote{\url{https://docs.jax.dev/en/latest/601/jaxpr.html}.}) before lowering to HLO. 
Starting from a root output type constraint---expressed in a domain-specific grammar bounding tensor rank, dimension sizes (e.g., multiples of $128$ to align with TPU lanes), and numeric data types---the generator samples a root operation and invokes its inverse type rule. 
Each operation's type rule defines a backward generator that samples valid operand shapes and types conditioned on the target output type, naturally handling non-injective rules (e.g., \texttt{conv}, \texttt{reshape}, or \texttt{abs} over complex inputs).
Sub-trees are recursively generated and stitched together using type-safe functional combinators (sequential composition, parallel concatenation, and lifting). 
Through configurable program templates, engineers can interleave deterministic graph skeletons with randomized operator subsets (e.g., restricting generation to $\{\texttt{slice}, \texttt{scatter}, \texttt{gather}\}$) and bound tree depth, comparing hardware execution against a reference evaluator to surface bug reproducers.

\subsubsection{Opportunistic Fuzzing on Real Graphs}
\label{sec:fleet_fuzzing}

While template-guided synthetic fuzzing effectively uncovers isolated pass defects, purely synthetic programs rarely capture the complex operator topologies or memory pressure of real-world models. Rather than relying solely on synthetic graphs, we extract the highest-impact fused operations directly from production workloads using fleet profiling telemetry. Our key insight is that production performance autotuning~\cite{cano2026catwild} is already an ideal fuzzer; that is, while searching for peak performance configurations, the autotuner systematically explores atypical parameters that stress-test the compiler. To turn performance tuning into continuous testing with zero net infrastructure replication, we embed tensor-output differential checks directly into our daily autotuning pipeline. 
As our autotuner evaluates candidate window/tile configurations on TPU hardware, it compares each candidate's output tensors against the compiler's default configuration within strict numerical tolerances---a process we further extend by permuting lowering-sensitive compiler flags across top fleet fusions. Together, these sweeps automatically surface both silent numerical deviations and runtime crashes. Crucially, fixing bugs in non-default candidate configurations proactively hardens the compiler: today's non-default window can become tomorrow's default as cost-model heuristics evolve, enabling our pipeline to intercept multiple severe defects before they reach production customers.

\subsubsection{Autonomous AI Agent Swarms for Fuzzing}
\label{sec:morris}

Repurposing fleet autotuning sweeps unlocked continuous verification over real workloads, but two bottlenecks remained: (1)~reaching compiler passes and IR layers outside the scope of standard autotuner knobs (such as raw LLO passes or custom kernel lowering), and (2)~the engineering toil required to triage, minimize, and patch discovered discrepancies. To scale both discovery and fixes, we run autonomous AI agent swarms in a closed loop on an internal agent harness built for large swarm orchestration and long horizon autonomous work.

The harness organizes a swarm as a tree of agent sessions: a lead agent spawns subagents, which may spawn their own, and a fleet runtime spreads them across workstations.
Each worker can receive an isolated checkout of the source tree, so concurrent patch attempts never collide.
Orchestration logic, such as fanning out over fuzzing targets or alternating verification and repair, is written as workflow scripts whose steps are journaled, so an interrupted run resumes where it stopped.
For long horizon work, a session pursues a goal until its completion gates (e.g., a test suite) pass, and durable watches poll state without spending model tokens, waking an agent only when a condition fires.
All agents share a TPU tool layer (ISA references, HLO and LLO dump inspection, layout and DMA descriptor decoding, single module recompilation, and profile analysis) and layered instructions that encode engineering policy, such as fix root causes, never disable tests, and take no externally visible action without human approval.

Rather than treating the compiler as a black-box, fuzzing subagents inspect XLA:TPU pass implementations to dynamically synthesize targeted fuzzer plugins. Organized by TPU architecture generation, the swarm targets complementary surfaces: end-to-end JAX/HLO compilation, direct LLO pass sequences, and per-pass graph invariant checks. To maximize throughput without exhausting physical accelerator quotas, fuzzing workers execute primarily on fast instruction-set simulators equipped with hardware-accurate MXU accumulator emulation, reserving physical TPU hardware for final bug confirmation.

A persistent challenge in generative fuzzing is distinguishing true compiler miscompilations from fuzzer generator bugs---inputs that pass structural verification but violate unchecked HLO semantic preconditions, triggering undefined behavior.  For example, generating a \texttt{reduce} with a non-neutral initial value (e.g., non-zero for a sum) yields false-positive discrepancies between the reference evaluator and TPU codegen.
Verification agents enforce a cross-pass semantic litmus test against formal HLO operation semantics; when a generator violation is detected, a subagent is automatically dispatched to constrain the fuzzer rather than filing a bogus compiler bug. 
Furthermore, for bugs discovered by fuzzing LLO passes directly, agents must construct an end-to-end HLO reproducer proving the faulty LLO pattern is genuinely reachable from high-level lowering.

For each confirmed miscompilation, patching subagents operating in isolated workspaces minimize the reproducer, audit candidate changes against hardware ISA specifications, and synthesize atomic compiler fixes constrained to a certain number of lines of code. 
To prioritize engineering attention, subagents quantify production severity by injecting targeted assertions into the faulty compiler path and compiling representative production model tiers (without launching full training jobs) to prove whether live workloads hit the defective code path. Every candidate patch, accompanying regression test (failing at main trunk, passing with the fix), and severity proof is then audited by an independent adversarial review swarm that checks numeric corner cases (e.g., IEEE-754 rounding boundaries, signedness modulo folding) and static analysis hygiene.

Throughout the workflow, both the human author and compiler maintainers remain strictly in the loop. Guided by the author, who steers target selection and triages complex findings, a coordinator agent queues verified patches and surfaces only the highest-severity fixes to avoid overwhelming maintainers or exhausting CI quotas. Once a change is routed for human code review, event-driven agents, built on the harness's watches, monitor review threads at zero idle token cost: they assist the author by dispatching worker agents to answer straightforward reviewer questions with \texttt{file:line} citations or apply requested edits, while immediately escalating open design questions to the human author and requiring explicit maintainer approval before any change can land.


\subsection{Hierarchical Isolation}
\label{sec:isolation}



When a large-scale computational graph miscompiles, end-to-end differential execution against a reference backend is unactionable; that is, comparing only the final model outputs reveals that a discrepancy occurred, but not which operation introduced it. 
At the same time, operator fusion merges long chains of operations into monolithic kernels whose intermediate values remain in VREGs and never materialize in off-chip memory, turning each fused kernel into an opaque black box during standard graph-level execution.
To bridge this gap between graph-level scale and kernel-level opacity, we localize miscompilations hierarchically in two stages. 
First, macro-graph isolation narrows the full graph down to a single failing subgraph or fusion by evaluating isolated sub-computations with constraint-propagated synthetic inputs. 
Second, once the culprit fusion is identified, micro-fusion inspection dives inside the intact fusion, capturing the intermediate VREG values of internal operations to pinpoint the exact instruction where hardware execution first diverges.

\subsubsection{Macro Graph Bisection}
\label{sec:macro_bisection}

To localize a numerical discrepancy within a large HLO graph, our macro-graph isolator (upstreamed to the OpenXLA project\footnote{\url{https://github.com/openxla/xla/blob/main/docs/hlo_isolation.md}}) decomposes the full graph into standalone subgraphs (primarily targeting element-wise and matmul fusions) while preserving physical tensor layouts and other configurations.
To make each extracted fusion a standalone, legal graph executable on a single TPU chip, the decomposer sanitizes entry parameters and outputs by inserting copy operations so that boundary tensors reside in off-chip HBM rather than on-chip VMEM, and updates VMEM scratchpad offsets for the target device. Deduplicated subgraphs are then deterministically sharded across parallel test workers. Each worker executes the fused subgraph on physical TPU hardware and compares its outputs against both an unfused version and a software reference evaluator\footnote{\url{https://github.com/openxla/xla/tree/main/xla/hlo/evaluator}.}. 
To minimize end-to-end triage time, the runner cascades through these reference checks and short-circuits as soon as any check passes, starting with a fast unfused hardware comparison before invoking the slower CPU-based reference evaluator.

Making offline subgraph isolation actionable in practice requires aggressively eliminating false positives, starting with input generation via reverse constraint propagation. Sampling naive random inputs for isolated subgraphs routinely triggers mathematical domain errors, such as negative operands to \texttt{sqrt}/\texttt{rsqrt}, zero divisors, or cascading overflows in large reductions, generating $\mathrm{NaN}$ and $\mathrm{Inf}$ values that both cause spurious mismatches ($\mathrm{NaN} \neq \mathrm{NaN}$) and mask real bugs behind saturation ($\mathrm{Inf} == \mathrm{Inf}$).
Rather than relying on heavyweight SMT solvers or slow rejection sampling, we employ a linear-time backward dataflow pass using interval arithmetic to propagate safe numerical bounds (scaled dynamically for downstream reductions) alongside structural and hardware alignment constraints. For subgraphs that still emit non-finite values, a static dataflow intent analyzer traces whether the $\mathrm{Inf}/\mathrm{NaN}$ outputs originate from intentional masking constants (e.g., $-\infty$ attention masks injected via \texttt{select}/\texttt{pad}\footnote{ \url{https://openxla.org/xla/operation_semantics}.}) without passing through unconstrained arithmetic operations.

Complementing domain-valid input synthesis, we hardened our counterfactual reference evaluator to eliminate benign hardware-software discrepancies. Specifically, when an isolated kernel only partially writes to an uninitialized scratchpad buffer, we automatically restrict output comparisons to the modified tensor region to avoid uninitialized buffer reads. Furthermore, the evaluator models nested TPU tiling layouts and emulates target-specific hardware numerics---including the Extended Unary Pipeline's (EUP) hardware approximations for transcendental functions (e.g., \texttt{exp}, \texttt{log}, \texttt{tanh}) and flushing tiny, near-zero subnormal numbers directly to zero---ensuring that remaining mismatches reflect genuine compiler bugs.

\subsubsection{Micro Fusion Inspection}
\label{sec:fusion_inspection}

While macro-graph isolation identifies which fusion diverged, it only observes the kernel's external HBM inputs and outputs.
Diving deeper to pinpoint the faulty instruction inside a fusion requires overcoming two systems hurdles. 
First, a classic pitfall in debugging fused TPU kernels is inserting barrier instructions to materialize and inspect intermediate tensors in HBM; because barriers split the fusion, they alter VMEM tiling, DMA pipelining, and LLO register allocation, frequently causing the bug under investigation to vanish (a compiler ``Heisenbug''). 
Second, dumping full intermediate tensors across a 10k-node graph or deep fusion chain rapidly exhausts off-chip HBM and host I/O bandwidth; instead, computing a compact 16-byte statistical summary ($\min$, $\max$, mean, and standard deviation) per tensor directly on-device allows our pipeline to monitor execution within existing memory budgets, reserving full tensor dumps strictly for the narrow window where divergence first appears.

To inspect a failing fusion without perturbing its generated code, our pipeline halts the TensorCore at the fusion entry, steps through the intact kernel, and reads the intermediate VREG values of each internal operation directly from hardware. Whenever an isolated fusion fails both the unfused TPU and reference evaluator checks, the isolator automatically triggers intra-fusion inspection: it re-evaluates the despecialized reference module with dynamic callbacks to populate an in-memory map of expected intermediate tensors, pairs reference and target instructions, and re-runs the intact fusion on hardware. 
As each internal instruction callback fires, runtime stepping enables both in-flight chunk-level debugging across stepped register chunks and full intermediate tensor reconstruction, reassembling captured register slices into the logical host tensor, comparing against the expected reference tensor, and immediately evicting it from memory.
On mismatch, the pipeline emits an interactive HTML report that highlights diverging internal instructions across nested fusions and embeds a tensor inspector displaying a 3D mismatch bounding box and peak error coordinates, enabling engineers to apply surgical per-instruction compiler metadata or fixes while keeping the fusion intact.


\section{Evaluation}

\begin{table*}[t]
\centering
\small
\begin{tabular}{@{}p{0.28\textwidth} r p{0.58\textwidth}@{}}
\toprule
\textbf{Compiler Subsystem} & \textbf{Defects} & \textbf{Primary Defect Categories} \\
\midrule
\textbf{TensorCore LLO \& Emitters} & $43\%$ & Sublane store mask corruption, unsigned comparison mismatches, narrow-integer unpack guards, multi-tile loop revisit and chunk-packing bugs \\
\addlinespace[2pt]
\textbf{HLO Simplifiers \& Rewrites} & $25\%$ & Nested dynamic-slice index overflow, degenerate scatter indexing, invalid loop trip-count and dilated-pad folding, space-to-batch halo underflow \\
\addlinespace[2pt]
\textbf{Precision \& Emulation Rewriters} & $14\%$ & Emulated $64$-bit integer divide-by-zero, IEEE-754 bitcast and narrowing cast omissions, sign-extension loss, complex trig overflow/underflow \\
\addlinespace[2pt]
\textbf{SparseCore Lowering} & $10\%$ & $2\text{D}$ interleaved float-to-int packing errors, sublane bit-reversal omissions, cross-core DMA synchronization and alignment faults \\
\addlinespace[2pt]
\textbf{Custom Kernel Lowering} & $8\%$ & Vector transpose op duplication, layout inference attribute mismatches, bitcast register-chunk corruption \\
\bottomrule
\end{tabular}
\caption{Distribution of compiler defects uncovered and resolved across five core XLA:TPU subsystems, illustrating how miscompilations concentrate in low-level backend emitters ($43\%$) and high-level algebraic simplifiers ($25\%$).}
\label{tab:bug_taxonomy}
\end{table*}


We evaluate our verification and triage methodology across Google's production XLA:TPU fleet, where it has already uncovered and driven the resolution of over \bugs{} compiler bugs (with \pendingbugs{} additional fixes currently underway).
While we also track and fix compile-time and runtime crashes, particularly those surfaced by our synthetic generator and fleet-graph fuzzing, the vast majority of our resolved bugs are silent numerical miscompilations, as we deliberately steered our autonomous agent swarms away from loud crashes to focus on silent data corruption (more on \S\ref{sec:steering_crash_corruption}).

We structure our evaluation around three questions: where in the compilation stack do silent numerical miscompilations reside? (\S\ref{sec:eval_taxonomy}); how deep and subtle are the miscompilations caught by our fuzzing pipeline? (\S\ref{sec:eval_depth}); and how effectively does hierarchical isolation identify true culprits? (\S\ref{sec:eval_isolation}).

\subsection{Bug Taxonomy Across the Compilation Stack}
\label{sec:eval_taxonomy}

To understand where compiler defects concentrate in practice, Table~\ref{tab:bug_taxonomy} breaks down the distribution of peer-reviewed bug fixes produced by our automated fuzzing campaigns across five core XLA:TPU subsystems.
Nearly half of all defects ($43\%$) reside in the lowest-level TensorCore LLO \& Emitters layer---arguably the most complex part of the compiler, where the compiler maps operations onto $8\times 128$ vector registers, sublane masks, and multi-dimensional VMEM tile loops. Upstream in the compilation pipeline high-level HLO Simplifiers ($25\%$) and Precision \& Emulation Rewriters ($14\%$) account for the next largest shares, while the remaining defects span specialized SparseCore Lowering ($10\%$) and Custom Kernel Lowering ($8\%$).

\subsection{Deep Miscompilations in the Wild}
\label{sec:eval_depth}

To illustrate the complexity of the defects in Table~\ref{tab:bug_taxonomy} and how deep in the compilation stack silent corruptions hide, we examine a few representative bugs.


\paragraph{Preempting Critical Regressions on Fleet Graphs}
Lowering fused subgraphs onto TPU on-chip memory requires orchestrating multi-dimensional tile loops that initialize VMEM accumulation buffers on the first visit to an output tile (window) and accumulate on subsequent revisits. Daily window-selection sweeps on fleet graphs caught some severe state-tracking bugs before customer escalations.
In one case, daily fleet fuzzing flagged a numerical divergence in fused \texttt{conv-reduce} kernels a few days before a critical production escalation where a customer's evaluation metrics dropped by $100\times$ after a compiler update fused a \texttt{convolution} with a \texttt{reduce} through a \texttt{bitcast}. Because a fused convolution dictates a multi-dimensional loop schedule that can revisit the same reduction output window across steps, the compiler allocates a tiny on-chip scalar state tracker to record visited windows. However, to save a few bytes of scalar scratchpad memory, a compile-time heuristic omitted this tracker whenever the loop bound of the reduced physical dimension was~$1$, ignoring that another convolution loop dimension (with bound~$2$) still revisited the same output tile and overwrote partial sums instead of accumulating them.
Although we had not yet landed the fix before the escalation hit, the silver lining was that seeing a fuzzer finding materialize as a production P0 reinforced the value of the effort across the team (\S\ref{sec:scalation}).
That lesson paid off soon after in a second fleet-fuzzer finding on windowed reductions with packed input chunks (packing factor $P > 1$), where the emitted loop check only initialized the VMEM accumulation buffer at loop index~$0$ rather than at every packed boundary ($P, 2P, \dots$); because we had prioritized a fix, it was already in flight when a customer nearest-neighbor pipeline reported silently losing $\sim\!30\%$ of top-ranked results, allowing us to restore top-$1000$ recall to $>99\%$ right away.

\paragraph{Agentic Discovery and Bug Fixing}
Listing~\ref{lst:agent_jax_repros} highlights four minimal JAX reproducers discovered and patched end-to-end by our autonomous agent swarm, illustrating both the subtlety of the bugs and the surgical fixes synthesized by the agents:
\textbf{(1)~Complex trigonometry (\texttt{bug1\_complex\_trig}):} Lowering \texttt{complex64} $\tan(x{+}yi)$ via $\sinh(y)\cosh(y)/(\cos^2 x + \sinh^2 y)$ overflowed $\sinh(y)$ to $\infty$ for $|y|{>}44.36$ (producing $\infty\times 0 = \mathrm{NaN}$), while in $\tanh(a{+}bi)$, a finite denominator exceeding $2^{126}$ ($|a|\in[43.67, 44.36]$) caused TPU \texttt{f32} reciprocal division ($1/\text{denom}$) to underflow to subnormal and flush to $0{+}0i$ instead of $\pm 1$. The agent fixed both by rewriting $\tan(z)$ via the identity $-i\tanh(-y{+}xi)$ and saturating the real part of $\tanh$ to $\operatorname{sign}(a)$ once $\text{denom} \ge 2^{126}$.
\textbf{(2)~Singleton-scatter vectorization (\texttt{bug2\_scatter\_int8}):} To vectorize a single-element scatter into a length-$N$ array, a TPU pass rewrites the scatter into a one-hot \texttt{select} mask by comparing the broadcasted index against an \texttt{iota} tensor $[0, 1, \dots, N{-}1]$. When the index was \texttt{int8} and $N > 128$, the pass emitted an \texttt{int8} \texttt{iota} that wrapped modulo $256$, causing a single write at index~$5$ to also overwrite index~$261$; the agent fixed this by widening narrow indices ($<32$-bit) to \texttt{int32} before generating the \texttt{iota} comparison.
\textbf{(3)~Small-batch Top-K (\texttt{bug3\_topk\_axis0}):} To parallelize an axis-$0$ Top-$K$ sort over a narrow $[N, B]$ matrix, the compiler folds axis~$0$ across the TPU's $128$ lanes into $[M, 128, B]$ ($M{=}N/128$) for a two-stage sort, where lane $l$ holds strided indices $m\cdot 128 + l$. However, index reconstruction mistakenly built a $[128, K, B]$ lane-offset \texttt{iota} along axis~$0$ and computed $\text{idx} + l\cdot M$ instead of $\text{idx}\cdot 128 + l$; the agent reshaped the offset \texttt{iota} to $[K, 128, B]$ (axis~$1$), scaled the first-stage index by $128$, and bypassed the strided split when \texttt{is\_stable=True}.
\textbf{(4)~Cum-sum constant folding (\texttt{bug4\_cumsum\_cast}):} An algebraic rewrite simplifies a cumulative sum over a broadcasted constant $c$ into $c \times [1, 2, \dots, N]$ by looking through unary \texttt{convert} chains. Because it only compared the initial constant's type (\texttt{f32}) against the final output type (\texttt{f32}), it skipped the narrowing \texttt{int32} truncation on $1.75$; the agent patched the pass to sequentially evaluate all intermediate \texttt{convert} types on the scalar literal before folding.

\definecolor{lstbg}{HTML}{F8FAFC}%
\definecolor{lstrule}{HTML}{CBD5E1}%
\definecolor{lstdiv}{HTML}{E2E8F0}%
\definecolor{lstkw}{HTML}{1E40AF}%
\definecolor{lstcmt}{HTML}{64748B}%
\definecolor{lstemph}{HTML}{0F766E}%
\definecolor{lstdec}{HTML}{6D28D9}%
\definecolor{lsthdr}{HTML}{0F172A}%
\definecolor{lstsub}{HTML}{475569}%
\definecolor{lsterr}{HTML}{B91C1C}%
\begin{lstlisting}[language=Python, basicstyle=\ttfamily\scriptsize\frenchspacing, columns=fixed, basewidth=0.53em, keepspaces=true, backgroundcolor=\color{lstbg}, frame=tb, framerule=0.6pt, rulecolor=\color{lstrule}, framesep=4pt, xleftmargin=2pt, xrightmargin=2pt, aboveskip=6pt, belowskip=4pt, keywordstyle=\color{lstkw}\bfseries, commentstyle=\color{lstcmt}\itshape, emph={jnp,lax,tan,tanh,broadcasted_iota,sort_key_val,broadcast_to,astype,cumsum,int8,int32,float32,at,set}, emphstyle=\color{lstemph}, escapeinside={(*@}{@*)}, caption={Example of minimal JAX reproducers for silent miscompilations discovered and patched by our AI agent swarm.}, label={lst:agent_jax_repros}, captionpos=b, float=htbp]
(*@\textbf{\textcolor{lsthdr}{\# (1) Complex trigonometry:}} \textcolor{lstsub}{tan NaN \& tanh 0-flush}@*)
(*@\textcolor{lstdec}{@jax.jit}@*)
def bug1_complex_trig(z1, z2):
  # z1=45j (*@$\to$@*) (*@\textcolor{lsterr}{\upshape NaN}@*); z2=44+0j (*@$\to$@*) (*@\textcolor{lsterr}{\upshape 0+0j}@*) (vs. 1j, 1+0j)
  return jnp.tan(z1), jnp.tanh(z2)
(*@\raisebox{0.6ex}{\rlap{\textcolor{lstdiv}{\rule{\linewidth}{0.4pt}}}}@*)
(*@\textbf{\textcolor{lsthdr}{\# (2) Singleton-scatter:}} \textcolor{lstsub}{int8 iota wraps at 128}@*)
(*@\textcolor{lstdec}{@jax.jit}@*)
def bug2_scatter_int8(x, idx, v):
  # x: f32[300], idx: int8(5) (*@$\to$@*) (*@\textcolor{lsterr}{\upshape writes x[5] \& x[261]}@*)
  return x.at[idx].set(v)
(*@\raisebox{0.6ex}{\rlap{\textcolor{lstdiv}{\rule{\linewidth}{0.4pt}}}}@*)
(*@\textbf{\textcolor{lsthdr}{\# (3) Small-batch Top-K:}} \textcolor{lstsub}{transposed lane iota}@*)
(*@\textcolor{lstdec}{@jax.jit}@*)
def bug3_topk_axis0(x):
  # x: f32[1024, 4] (axis 0) (*@$\to$@*) (*@\textcolor{lsterr}{\upshape top-k indices corrupted}@*)
  idx = lax.broadcasted_iota(jnp.int32, x.shape, dim=0)
  v, i = lax.sort_key_val(x, idx, dim=0, is_stable=False)
  return v[:8], i[:8]
(*@\raisebox{0.6ex}{\rlap{\textcolor{lstdiv}{\rule{\linewidth}{0.4pt}}}}@*)
(*@\textbf{\textcolor{lsthdr}{\# (4) Cum-sum constant folding:}} \textcolor{lstsub}{drops int32 cast}@*)
(*@\textcolor{lstdec}{@jax.jit}@*)
def bug4_cumsum_cast():
  # Returns (*@\textcolor{lsterr}{\upshape [1.75, 3.5, ...]}@*) instead of [1., 2., ...]
  x = jnp.broadcast_to(1.75, (4,))
  x = x.astype(jnp.int32).astype(jnp.float32)
  return jnp.cumsum(x)
\end{lstlisting}

\begin{figure*}[t]
\centering
\begin{minipage}[t]{0.32\textwidth}
{\small\textbf{(a) High-Level Code}}\par
\scriptsize
\begin{verbatim}
// QAT / multi-word residual
q = x.astype(f16)
r = x - q.astype(f32)
\end{verbatim}
\end{minipage}\hfill
\begin{minipage}[t]{0.34\textwidth}
{\small\textbf{(b) Optimized Fused HLO}}\par
\scriptsize
\begin{verbatim}
// Both converts remain in HLO
%q = f16[N] convert(%x)
%u = f32[N] convert(%q)
ROOT %r = f32[N] subtract(%x, %u)
\end{verbatim}
\end{minipage}\hfill
\begin{minipage}[t]{0.32\textwidth}
{\small\textbf{(c) Lowered Code}}\par
\scriptsize
\begin{verbatim}
// Casts skipped in registers
v0 = vload(x)    // f32 VREG
v1 = v0; v2 = v1 // casts skipped
v3 = v0 - v2     // == 0.0
\end{verbatim}
\end{minipage}
\caption{A silent numerical discrepancy introduced when lowering a fused kernel.
The optimized HLO \textbf{(b)} preserves the user's explicit narrowing and widening casts \textbf{(a)}, but the lowered code \textbf{(c)} keeps intermediate values in 32-bit vector registers and skips the \texttt{f16} round-trip, causing the subtraction to evaluate $x - x = 0.0$.}
\label{fig:silent_miscompilation_examples}
\end{figure*}

\subsection{Hierarchical Isolation and Defeating Heisenbugs}
\label{sec:eval_isolation}

Figure~\ref{fig:silent_miscompilation_examples} illustrates why silent numerical discrepancies inside fusions are difficult to diagnose on TPUs: the optimized HLO IR looks right, and the divergence is introduced only when lowering operations inside the fused kernel.
In quantization-aware training (QAT) and multi-word arithmetic schemes, user code casts an \texttt{f32} tensor down to a narrow type (\texttt{f16}, \texttt{bf16}, or \texttt{fp8}) and back to \texttt{f32} so that $r = x - \mathrm{f32}(\mathrm{f16}(x))$ captures the rounding residual (\ref{fig:silent_miscompilation_examples}a).
Both casts stay intact in the optimized fused HLO (\ref{fig:silent_miscompilation_examples}b), but XLA's default precision policy keeps intermediates in 32-bit \textsc{VREG}s and skips both casts (\texttt{v1 = v0; v2 = v1} in Figure~\ref{fig:silent_miscompilation_examples}c), so the subtraction evaluates $x - x = 0.0$ via self-cancellation ($0.0\%$ nonzero elements on TPU vs.\ $93.7$--$100\%$ on the reference platform).
When this fusion is embedded inside a multi-layer Transformer block, traditional unfused debugging fails to reproduce the bug. Splitting the fusion spills intermediate tensors to 16-bit HBM, accidentally restoring the rounding and matching the reference bit-for-bit.

Running our hierarchical isolation pipeline on the intact block pinpointed the root cause in two steps: macro-graph isolation ranked candidates by maximum relative error to immediately surface the QAT fusion ($100\%$ relative error, with all $131{,}072$ elements collapsed to $0.0$) ahead of minor floating-point drift in surrounding matrix multiplications, and micro-fusion inspection confirmed inside the intact kernel that both \texttt{convert} ops left the input unrounded (VREG $\equiv x$) before the root \texttt{subtract} canceled them to $0.0$. Enforcing explicit 16-bit register rounding on that fusion alone restored all nonzero QAT residuals bit-for-bit ($0.00\%$ error) without affecting 32-bit accumulation elsewhere in the model.

\section{Experiences and Lessons Learned}
\label{sec:lessons}

In this section, we describe the key lessons learned from deploying our XLA:TPU verification pipeline at scale, spanning fuzzing strategy, developer adoption, and human-AI compiler workflows.

\subsection{Compiler Fuzzing}

\subsubsection{Performance Autotuners Are Natural Fuzzers}
In compiler engineering, performance tuning and correctness verification are traditionally treated as separate workflows with distinct infrastructure.
In practice, however, because autotuners aggressively explore cases in pursuit of faster programs, their search space overlaps heavily with the boundary conditions where subtle miscompilations hide.
Recognizing this synergy allowed us to avoid building costly fuzzing infrastructure from scratch, and instead evolve our production autotuning system~\cite{cano2026catwild} into a dual-purpose engine that simultaneously drives efficiency and continuous, fleet-wide correctness verification at near-zero marginal infrastructure cost.

\subsubsection{Synthetic Breadth Meets Production Depth}
Neither synthetic fuzzing nor production trace mining is sufficient in isolation. Template-guided synthetic generators (\S\ref{sec:xlasmith}) excel at probing newly developed passes and structural corner cases with compact, fast-running graphs; however, unconstrained synthetic fuzzers can easily stall in early validation checks or spend cycles exercising cold code paths. Developing both flagship ML workloads and the compiler stack under the same roof placed us in a unique position to complement synthetic generation with real fleet-mined graphs, ensuring our verification also targets the exact operator topologies, memory pressures, and codegen hot paths that dominate production.

\subsubsection{Fixing Latent Debt vs. Hardening a Moving Target}
Our pipeline simultaneously served two roles: addressing latent technical debt across existing compiler passes and proactively guarding against future regressions as new TPU generations and model architectures emerge.
At the same time, operating at hyperscale reinforced a humbling reality: across the vast configuration space of a modern ML compiler, correctness is not a static milestone to reach, but an ongoing operational discipline.
Rather than viewing verification as a one-time static audit, embedding continuous differential checks and rapid isolation into daily fleet operations transforms correctness from a reactive firefight into a sustainable, ongoing engineering discipline.

\subsection{Developer Adoption}

\subsubsection{Overcoming Developer Skepticism}
\label{sec:scalation}
Compiler engineers are naturally skeptical of bugs triggered by synthetic corner cases or non-default compiler configurations, often questioning whether they can occur in real models.
In practice, assessing the severity of a fuzzer finding upfront is genuinely difficult; however, seeing bugs initially flagged by our fuzzers in non-default window configurations later surface as production escalations underscored that latent corner cases do matter as workloads and cost models evolve.
Finally, because engineers rarely act on raw fuzzer logs, automatically translating a failing configuration or bisected subgraph into a standalone, hermetic unit test is the difference between an ignored alert and a merged fix.
This investment paid a second dividend: a fast, self-contained failing unit test provides the exact executable specification required to point an autonomous AI agent at a bug to synthesize and verify a fix.

\subsubsection{False-Positive Fatigue}
In our early fuzzing runs, naive differential testing generated far more testing artifacts than true compiler bugs, stemming from out-of-domain random inputs (e.g., negative operands to \texttt{rsqrt}), uninitialized VMEM padding, and benign floating-point non-associativity. Even a 5\% false-positive rate is fatal to developer adoption; once compiler engineers spend hours triaging a ``miscompilation'' only to discover a testing artifact, they quickly tune out future alerts. Achieving high signal-to-noise required treating every false positive with the same rigor as a compiler bug, investing as much engineering effort into reverse constraint propagation and reference-evaluator fidelity as into the fuzzers themselves.

\subsection{Human-Centric AI Agent Workflows}

\subsubsection{Steering Agents from Loud Crashes to Silent Corruptions}
\label{sec:steering_crash_corruption}
Autonomous AI agents proved remarkably effective at both uncovering compiler bugs and synthesizing verified fixes across the stack.
However, unconstrained fuzzing agents naturally gravitate toward the easiest failure signal, i.e., loud compile-time assertion failures and unhandled edge-case crashes, which can quickly saturate human review bandwidth with low-severity fixes. Maximizing engineering impact required deliberately steering our agent swarms away from loud crashes and toward silent numerical miscompilations, where an agent's ability to exhaustively audit index arithmetic, floating-point corner cases, and hardware lane mappings delivers the highest return on reviewer time.

\subsubsection{Reviewer Burden Scales Non-Linearly with Fix Complexity}
When deploying AI agents to draft compiler fixes, unconstrained patches often touch hundreds of lines across multiple passes, imposing a steep cognitive burden on human maintainers. 
Counterintuitively, forcing agents to emit small, atomic patches, even when a broader refactoring might ultimately be preferable, proved vastly superior. A compact patch paired with a minimal failing test allows a human expert to verify in seconds that the underlying miscompilation is real; from there, the maintainer can either accept the surgical fix, refine it directly, or instruct the agent to explore an alternative design.

\subsubsection{Adversarial Pre-Filtering and Opt-in Reviews}

Autonomous bug-fixing pipelines can also easily overwhelm a compiler team if every plausible patch is mailed directly to maintainers. 
Gatekeeping outgoing changes behind an independent adversarial review swarm, strict severity ranking, and an opt-in human reviewer model ensured that human engineers only saw high-confidence, pre-audited fixes, preserving engineering goodwill.

\subsubsection{Design Invariants in AI-Assisted Compiler Code}
With clear specifications and human review, modern models excel at local implementation hygiene---rarely introducing off-by-one indexing, type-narrowing, or boundary-check errors---and instead fail mainly on undocumented cross-pass assumptions or hardware contracts that exist only in a maintainer's mental model. Scaling reliable AI-assisted compiler engineering thus requires turning these implicit mental contracts into explicit IR verifiers, pass preconditions, and machine-checkable assertions.

\section{Related Work}

We discuss related work in the areas of differential testing and compiler fuzzing, formal verification of compilers and rewrites, and AI-driven compiler bug finding. 

\paragraph{Differential Testing and Compiler Fuzzing:}
Prior work synthesizes scalar programs~\cite{yang2011finding, livinskii2020yarpgen} or cross-runtime differential tests~\cite{pham2019cradle, gu2022muffin}, while ML compiler fuzzers generate computational DAGs~\cite{10.1145/3575693.3575707, 10.1145/3597926.3598105, 10.1145/3611643.3616337}, rewrite e-graphs~\cite{10.1145/3689757}, or mutate compiler IRs~\cite{10.1145/3597926.3598053, liu2022tzer, 10.1109/ASE56229.2023.00120, 10.1109/ICSE55347.2025.00037, 10.1145/3763161}. However, small synthetic graphs miss whole-program resource constraints. We instead pair synthetic fuzzing with full-scale fleet traces, using dataflow constraint propagation to suppress floating-point false positives and localize bugs.

\paragraph{Formal Verification of Compilers and Tensor Rewrites:}
Beyond scalar verification~\cite{10.1145/1538788.1538814, 10.1145/3453483.3454030}, recent work verifies ML graph rewrites using SMT solvers~\cite{10.1145/3704865} or interactive theorem provers like \textsc{Lean~4} for verifying compiler code generation~\cite{10.1145/3819802.3820579}. 
Instead of proving isolated rewrites statically, our work validates stateful, multi-pass lowering (HLO $\rightarrow$ LLO $\rightarrow$ VLIW) on real hardware and autonomously patches defects.

\paragraph{AI-Guided Compiler Bug Finding:}
Recent frameworks leverage LLMs to synthesize bug-triggering inputs or localize miscompilations from diagnostic traces~\cite{10.1145/3689736, 10.1145/3597503.3639121, lebar2026}. Instead, we combine fleet-scale autotuning with multi-agent swarms that close the entire loop, from codebase-aware multi-layer fuzzing and semantic filtering to ISA-audited patch synthesis and adversarial review.

\section{Conclusions}

As modern ML compilers rely on increasingly aggressive optimizations to saturate custom accelerators, silent numerical miscompilations pose a first-order risk to large-scale training and serving. In this paper, we presented a unified reliability framework for XLA:TPU combining template-guided synthetic fuzzing, differential verification piggybacked on production autotuning, and hierarchical graph-to-kernel isolation, amplified by human-in-the-loop AI agent swarms. Deployed at datacenter scale, our methodology has resolved over~\bugs{} compiler bugs (\pendingbugs{} more underway), offering a practical blueprint for hardening production ML compilers without sacrificing optimization velocity.


\section*{Acknowledgements}
\vspace{-0.5em}
We thank Mike Burrows for his valuable and constructive feedback on earlier drafts of this manuscript. We are also grateful to the many collaborators and contributors who supported and advanced the XLA:TPU correctness effort over the years.


\bibliography{biblio}
\bibliographystyle{mlsys2025}



\end{document}